\documentclass[letterpaper, 10 pt, conference]{ieeeconf}  

\IEEEoverridecommandlockouts                              

\usepackage{amsmath} 
\usepackage{amssymb}  
\usepackage{mathtools}
\usepackage{siunitx}

\usepackage[table,xcdraw]{xcolor}
\usepackage{graphicx}
\usepackage{tabularx,booktabs}
\newcolumntype{Y}{>{\centering\arraybackslash}X}

\usepackage{todonotes}
\usepackage[nolist]{acronym}
\usepackage{flushend}
\usepackage[font=small]{caption}
\usepackage[font=small]{subcaption}

\makeatletter
\renewcommand*\env@matrix[1][*\c@MaxMatrixCols c]{%
	\hskip -\arraycolsep
	\let\@ifnextchar\new@ifnextchar
	\array{#1}}
\makeatother

\makeatletter
\let\NAT@parse\undefined
\makeatother
\usepackage{hyperref}  

\begin{acronym}
\acro{GPU}{Graphics Processing Units}
\acro{DNN}{Deep Neural Networks}
\acro{CNN}{Convolutional Neural Networks}
\acro{RNN}{Recurrent Neural Networks}
\acro{GNN}{Graph Neural Networks}
\acro{GAT}{Graph Attention Network}
\acro{GCN}{Graph Convolutional Networks}
\acro{GRU}{Gated Recurrent Unit}
\acro{LSTM}{Long Short-Term Memory}
\acro{KF}{Kalman Filters}
\acro{BEV}{Bird's Eye View}
\acro{GAN}{Generative Adversarial Networks}
\acro{MTP}{Multi-Modal Trajectory Prediction}
\acro{MLP}{Multilayer Perceptron}
\acro{MAE}{Mean Absolute Error}
\acro{MSE}{Mean Squared Error}
\acro{ADE}{Average Displacement Error}
\acro{FDE}{Final Displacement Error}
\acro{DKM}{Deep Kinematic Models}
\acro{FFW-ASP}{Feed-Forward Action-Space Prediction}
\acro{SSP-ASP}{Self-Supervised Action-Space Prediction}
\acro{VAE}{Variational Autoencoder}
\acro{CVAE}{Conditional Variational Autoencoder}
\acro{AV}{autonomous vehicle}
\acro{MHA}{Multi-Head Attention}
\end{acronym}

\title{\LARGE \bf
StarNet: Joint Action-Space Prediction with Star Graphs and \linebreak Implicit Global Frame Self-Attention
}
\author{Faris Janjo\v{s}$^{1}$, Maxim Dolgov$^{1}$, and J. Marius Z\"ollner$^{2}$
	\thanks{$^{1}$ Robert Bosch GmbH, Corporate Research, Advanced Autonomous Systems, 71272 Renningen, Germany. {\tt\small \{faris.janjos, maxim.dolgov\}@de.bosch.com}}%
	\thanks{$^{2}$ Research Center for Information Technology (FZI), 76131 Karlsruhe, Germany.
		{\tt\small zoellner@fzi.de}}%
	\thanks{{This work was financially supported by the Federal Ministry of Economic Affairs and Energy of Germany, grant number 19A20026H, based on a decision of the German Bundestag.}}
}

\begin{document}

\maketitle
\thispagestyle{empty}
\pagestyle{empty}

\begin{abstract}
In this work, we present a novel multi-modal multi-agent trajectory prediction architecture, focusing on map and interaction modeling using graph representation. For the purposes of map modeling, we capture rich topological structure into vector-based star graphs, which enable an agent to directly attend to relevant regions along polylines that are used to represent the map. We denote this architecture StarNet, and integrate it in a single-agent prediction setting. As the main result, we extend this architecture to joint scene-level prediction, which produces multiple agents' predictions simultaneously. The key idea in joint-StarNet is integrating the awareness of one agent in its own reference frame with how it is perceived from the points of view of other agents. We achieve this via masked self-attention. Both proposed architectures are built on top of the action-space prediction framework introduced in our previous work, which ensures kinematically feasible trajectory predictions. We evaluate the methods on the interaction-rich inD and INTERACTION datasets, with both StarNet and joint-StarNet achieving improvements over state of the art.
\end{abstract}
\section{INTRODUCTION}\label{sec:intro}
Accurate prediction of the driving situation is a major cornerstone for achieving performant full autonomy of self-driving cars. Despite a strong research and industry focus, there are many problems to be solved, such as understanding complex social interactions among different agents and effectively incorporating rich topological information. Other important aspects are prediction of multi-modal trajectories, conditioning predictions on assumed goals of given agents, as well as achieving reasonable long-term predictions. In tackling these challenges, \ac{DNN} have shown great results over classical robotics approaches, especially for the use-case of urban driving.

The challenges of environment representation and interaction modeling are tightly coupled, e.g. the maneuvers of two negotiating vehicles in a highly-interactive situation are constrained by the road topology. Therefore, a learned model must consider the effects of topology on the driving situation. This makes the representation of map information paramount: it should enable explicitly relating to map elements such as lane centerlines and road boundaries, as well as segments along these elements. Furthermore, it must allow the model to discern between more and less important segments. In this sense, a direct, explicit representation of map geometry serves to condition the social interaction.

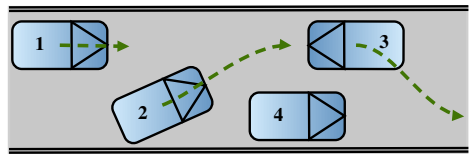
\begin{figure}
	\hspace{-25pt}
	\centering
	\scalebox{0.6}{

\tikzset {_72opxpz85/.code = {\pgfsetadditionalshadetransform{ \pgftransformshift{\pgfpoint{0 bp } { 0 bp }  }  \pgftransformrotate{0 }  \pgftransformscale{2 }  }}}
\pgfdeclarehorizontalshading{_obdltjlrn}{150bp}{rgb(0bp)=(0.81,0.91,0.98);
	rgb(37.5bp)=(0.81,0.91,0.98);
	rgb(62.5bp)=(0.39,0.58,0.76);
	rgb(100bp)=(0.39,0.58,0.76)}


\tikzset {_ds7q142ek/.code = {\pgfsetadditionalshadetransform{ \pgftransformshift{\pgfpoint{0 bp } { 0 bp }  }  \pgftransformrotate{0 }  \pgftransformscale{2 }  }}}
\pgfdeclarehorizontalshading{_hflgux5lv}{150bp}{rgb(0bp)=(0.81,0.91,0.98);
	rgb(37.5bp)=(0.81,0.91,0.98);
	rgb(62.5bp)=(0.39,0.58,0.76);
	rgb(100bp)=(0.39,0.58,0.76)}


\tikzset {_ckhuspsc7/.code = {\pgfsetadditionalshadetransform{ \pgftransformshift{\pgfpoint{0 bp } { 0 bp }  }  \pgftransformrotate{0 }  \pgftransformscale{2 }  }}}
\pgfdeclarehorizontalshading{_0isxx0obf}{150bp}{rgb(0bp)=(0.81,0.91,0.98);
	rgb(37.5bp)=(0.81,0.91,0.98);
	rgb(62.5bp)=(0.39,0.58,0.76);
	rgb(100bp)=(0.39,0.58,0.76)}


\tikzset {_rircv0a7z/.code = {\pgfsetadditionalshadetransform{ \pgftransformshift{\pgfpoint{0 bp } { 0 bp }  }  \pgftransformrotate{-180 }  \pgftransformscale{2 }  }}}
\pgfdeclarehorizontalshading{_ambx4rgqd}{150bp}{rgb(0bp)=(0.81,0.91,0.98);
	rgb(37.5bp)=(0.81,0.91,0.98);
	rgb(62.5bp)=(0.39,0.58,0.76);
	rgb(100bp)=(0.39,0.58,0.76)}
\tikzset{every picture/.style={line width=0.75pt}} 

\begin{tikzpicture}[x=0.75pt,y=0.75pt,yscale=-1,xscale=1]
	
	\draw  [color={rgb, 255:red, 0; green, 0; blue, 0 }  ,draw opacity=0.19 ][fill={rgb, 255:red, 128; green, 128; blue, 128 }  ,fill opacity=0.43 ][line width=1.5]  (2,4.5) -- (387.18,4.5) -- (387.18,127.08) -- (2,127.08) -- cycle ;
	\path  [shading=_obdltjlrn,_72opxpz85] (4.8,24.6) .. controls (4.8,20.18) and (8.38,16.6) .. (12.8,16.6) -- (76.66,16.6) .. controls (81.08,16.6) and (84.66,20.18) .. (84.66,24.6) -- (84.66,48.6) .. controls (84.66,53.02) and (81.08,56.6) .. (76.66,56.6) -- (12.8,56.6) .. controls (8.38,56.6) and (4.8,53.02) .. (4.8,48.6) -- cycle ; 
	\draw  [line width=1.5]  (4.8,24.6) .. controls (4.8,20.18) and (8.38,16.6) .. (12.8,16.6) -- (76.66,16.6) .. controls (81.08,16.6) and (84.66,20.18) .. (84.66,24.6) -- (84.66,48.6) .. controls (84.66,53.02) and (81.08,56.6) .. (76.66,56.6) -- (12.8,56.6) .. controls (8.38,56.6) and (4.8,53.02) .. (4.8,48.6) -- cycle ; 
	
	\draw  [line width=1.5]  (84.57,36.3) -- (54.76,55.88) -- (54.57,17) -- cycle ;
	\path  [shading=_hflgux5lv,_ds7q142ek] (89.58,92.31) .. controls (87.78,88.28) and (89.59,83.55) .. (93.62,81.75) -- (151.94,55.72) .. controls (155.97,53.92) and (160.7,55.73) .. (162.5,59.77) -- (172.28,81.69) .. controls (174.08,85.72) and (172.27,90.45) .. (168.24,92.25) -- (109.92,118.28) .. controls (105.89,120.08) and (101.16,118.27) .. (99.36,114.23) -- cycle ; 
	\draw  [line width=1.5]  (89.58,92.31) .. controls (87.78,88.28) and (89.59,83.55) .. (93.62,81.75) -- (151.94,55.72) .. controls (155.97,53.92) and (160.7,55.73) .. (162.5,59.77) -- (172.28,81.69) .. controls (174.08,85.72) and (172.27,90.45) .. (168.24,92.25) -- (109.92,118.28) .. controls (105.89,120.08) and (101.16,118.27) .. (99.36,114.23) -- cycle ; 
	
	\draw  [line width=1.5]  (167.19,70.49) -- (147.94,100.52) -- (131.93,65.09) -- cycle ;
	\path  [shading=_0isxx0obf,_ckhuspsc7] (204.4,84.4) .. controls (204.4,79.98) and (207.98,76.4) .. (212.4,76.4) -- (276.26,76.4) .. controls (280.68,76.4) and (284.26,79.98) .. (284.26,84.4) -- (284.26,108.4) .. controls (284.26,112.82) and (280.68,116.4) .. (276.26,116.4) -- (212.4,116.4) .. controls (207.98,116.4) and (204.4,112.82) .. (204.4,108.4) -- cycle ; 
	\draw  [line width=1.5]  (204.4,84.4) .. controls (204.4,79.98) and (207.98,76.4) .. (212.4,76.4) -- (276.26,76.4) .. controls (280.68,76.4) and (284.26,79.98) .. (284.26,84.4) -- (284.26,108.4) .. controls (284.26,112.82) and (280.68,116.4) .. (276.26,116.4) -- (212.4,116.4) .. controls (207.98,116.4) and (204.4,112.82) .. (204.4,108.4) -- cycle ; 
	
	\draw  [line width=1.5]  (284.17,96.1) -- (254.36,115.68) -- (254.17,76.8) -- cycle ;
	\path  [shading=_ambx4rgqd,_rircv0a7z] (333.63,48.68) .. controls (333.63,53.1) and (330.04,56.68) .. (325.62,56.67) -- (261.76,56.53) .. controls (257.34,56.52) and (253.77,52.93) .. (253.77,48.52) -- (253.83,24.52) .. controls (253.83,20.1) and (257.42,16.52) .. (261.84,16.53) -- (325.7,16.67) .. controls (330.12,16.68) and (333.69,20.27) .. (333.69,24.68) -- cycle ; 
	\draw  [line width=1.5]  (333.63,48.68) .. controls (333.63,53.1) and (330.04,56.68) .. (325.62,56.67) -- (261.76,56.53) .. controls (257.34,56.52) and (253.77,52.93) .. (253.77,48.52) -- (253.83,24.52) .. controls (253.83,20.1) and (257.42,16.52) .. (261.84,16.53) -- (325.7,16.67) .. controls (330.12,16.68) and (333.69,20.27) .. (333.69,24.68) -- cycle ; 
	
	\draw  [line width=1.5]  (253.89,36.82) -- (283.74,17.3) -- (283.85,56.18) -- cycle ;
	\draw [line width=1.5]    (2,124) -- (388.18,123.08)(2,127) -- (388.19,126.08) ;
	\draw [color={rgb, 255:red, 65; green, 117; blue, 5 }  ,draw opacity=1 ][line width=2.25]  [dash pattern={on 6.75pt off 4.5pt}]  (44.73,36.6) -- (98.18,37.5) ;
	\draw [shift={(103.18,37.58)}, rotate = 180.96] [fill={rgb, 255:red, 65; green, 117; blue, 5 }  ,fill opacity=1 ][line width=0.08]  [draw opacity=0] (10,-4.8) -- (0,0) -- (10,4.8) -- cycle    ;
	\draw [color={rgb, 255:red, 65; green, 117; blue, 5 }  ,draw opacity=1 ][line width=2.25]  [dash pattern={on 6.75pt off 4.5pt}]  (130.93,87) .. controls (158.7,76.74) and (199.94,37.5) .. (234.61,36.78) ;
	\draw [shift={(239.46,36.94)}, rotate = 185.14] [fill={rgb, 255:red, 65; green, 117; blue, 5 }  ,fill opacity=1 ][line width=0.08]  [draw opacity=0] (10,-4.8) -- (0,0) -- (10,4.8) -- cycle    ;
	\draw [color={rgb, 255:red, 65; green, 117; blue, 5 }  ,draw opacity=1 ][line width=2.25]  [dash pattern={on 6.75pt off 4.5pt}]  (293.73,36.6) .. controls (324.8,36.54) and (346.99,88.34) .. (379.93,95.79) ;
	\draw [shift={(384.66,96.54)}, rotate = 185.14] [fill={rgb, 255:red, 65; green, 117; blue, 5 }  ,fill opacity=1 ][line width=0.08]  [draw opacity=0] (10,-4.8) -- (0,0) -- (10,4.8) -- cycle    ;
	\draw [line width=1.5]    (2,4.77) -- (388.18,3.85)(2,7.77) -- (388.19,6.85) ;
	
	\draw (22.5,28) node [anchor=north west][inner sep=0.75pt]  [font=\large] [align=left] {\textbf{1}};
	\draw (109,86) node [anchor=north west][inner sep=0.75pt]  [font=\large] [align=left] {\textbf{2}};
	\draw (312,26.5) node [anchor=north west][inner sep=0.75pt]  [font=\large] [align=left] {\textbf{3}};
	\draw (223,88) node [anchor=north west][inner sep=0.75pt]  [font=\large] [align=left] {\textbf{4}};

\end{tikzpicture}}
	\caption{\small Example of a highly interactive situation requiring joint scene prediction: vehicle 1 slows down to allow vehicle 2 to overtake parked vehicle 4; vehicle 3 must reverse behind vehicle 4.}
	\label{fig:interaction_example}
	\vspace{-16pt}
\end{figure}

Predictions for interacting vehicles in a driving scene must be consistent. Therefore, it is desirable to do a joint prediction and to consider mutual social interaction (example in Fig. \ref{fig:interaction_example}). In contrast, predicting separately for each agent assumes that each vehicle considers others as part of its local environment. In this way, the scene at hand is considered from the point of view of each of the vehicles while marginalizing other vehicles, which is redundant and may lead to inconsistent predictions. This brings the question, how can we share this mutual local information? A potential solution is predicting jointly in a global reference frame; this eliminates the redundancy, but at the cost of an arbitrary dependency on the origin placement. Otherwise, performing individual, marginal predictions hinders sharing of intention information between local representations of the same agent.

In handling the joint nature of the prediction problem, many works use global representations centered around the~\ac{AV}~\cite{casas2020spagnn}, \cite{casas2020implicit}, \cite{cui2021lookout}, \cite{suo2021trafficsim}, \cite{scibior2021imagining}, in the form of sensor data or generated \ac{BEV} images. In order to obtain agent-specific features, they extract patches around the agents of interest, derive corresponding latent information, and perform individual or joint predictions, provided that the latent features are combined. This introduces twofold disadvantages. First, grid-based environment representation assumes that the model extracts map objects and agents implicitly from individual pixels while inferring semantic knowledge. And second, the networks are tasked with associating information about the same agents from different sets of latent features, which might vary if the agents' perspectives are different. In particular, if there is no overlap among the initial patches, it might be difficult to infer relational information between two agents if they don't exist in each other's immediate environment representations.
 
In our approach, we help our models learn by explicitly relating the same agents from other agents' perspectives. Furthermore, we enable them to learn to attend to the most relevant map elements and their segments directly. The contributions of our work are the following:
\begin{itemize}
\item A novel, star-graph-of-vectors polyline representation, with unconstrained field-of-view (given a map description) and direct modeling of most relevant segments.
\item A mechanism to explicitly combine features from multiple reference frames via self-attention~\cite{vaswani2017attention} while masking out irrelevant information, enabling joint scene-level prediction.
\item State-of-the-art performance on the INTERACTION dataset~\cite{zhan2019interaction}, containing challenging roundabout, intersection, and highway merge scenarios.
\end{itemize}
\section{RELATED WORK}
The field of trajectory prediction has generated an exhaustive literature~\cite{rudenko2020human}. Within the field of autonomous driving, the multifaceted nature of the problem yields works that focus on specific challenges within the overall landscape. They include but are not limited to: environment representation~\cite{gao2020vectornet}, \cite{liang2020learning}, \cite{hu2020scenario}, multi-agent interaction~\cite{tang2019multiple}, \cite{li2021rain}, \cite{tolstaya2021identifying}, multi-modality~ \cite{chai2019multipath}, \cite{zhao2020tnt}, \cite{liu2021multimodal}, goal-conditioning~\cite{rhinehart2019precog}, \cite{khandelwal2020if}, \cite{ngiam2021scene}, and kinematic constraints~\cite{cui2019deep}, \cite{phan2020covernet}, \cite{janjos2021action}. Furthermore, some works integrate prediction with detection and planning~\cite{zeng2019end}, \cite{zeng2020dsdnet}, \cite{cui2021lookout}, as an important step in achieving end-to-end self-driving. Our work focuses on \textbf{map representation} and \textbf{joint interaction modeling}. In addressing these challenges, approaches using~\ac{CNN}~\cite{lecun2015deep} and~\ac{GNN}~\cite{wu2020comprehensive} are prevalent, with \ac{GNN}s increasingly used over \ac{CNN}s.

\subsection{Graph-based map representation}
Encoding map information into graphs and using \ac{GNN}s offers several advantages over image-based inputs used in conjunction with \ac{CNN}s. \ac{CNN}s extract locational features from rasterized \ac{BEV} images or grids with LiDAR point clouds. In contrast, \ac{GNN}s can learn directly from graph-based representations, which encode the geometric structure of the road into nodes and edges. In doing so, they alleviate the need to infer objects from pixels, as well as improve efficiency due to fewer weights. Furthermore, graphs benefit from a larger field of view than rasters, which are usually restricted by image dimensions and resolution. These factors contribute to a substantially higher representation density.

VectorNet~\cite{gao2020vectornet} is a seminal work that uses graph-based map representations. This approach fits polylines to map elements and dissects them into their constituent vectors. Then, fully-connected graphs are constructed per map element, which are aggregated by a \ac{GNN} into a feature vector. This procedure is used as a basis in further works~\cite{zhao2020tnt}, \cite{tolstaya2021identifying}, \cite{liu2021multimodal}. Alternative approaches are~\cite{liang2020learning} and~\cite{zeng2021lanercnn}, redefining the graph convolution operation with additional adjacency matrices in order to capture a larger receptive field. Consequently, they are able to capture a longer range longitudinally, as well as account for the different semantic meaning of lateral lanes. Another class of works is~\cite{pan2020lane} and~\cite{khandelwal2020if}, which model attention to specific lanes or sections along a reference polyline, constructed by concatenating individual polyline points. In the case of~\cite{khandelwal2020if}, this enables placement of hypothetical goals along the polyline to condition the trajectory prediction.

Works like VectorNet construct fully-connected subgraphs for each map element in order to mitigate the \ac{GNN} information bottleneck\footnote{Modeling the full map connectivity as a mesh-like natural graph and interweaving the agents nodes would result in long chains of propagated information. Learning over such a graph would necessitate many iterations of message passing and yield over-smoothed node embeddings.}~\cite{alon2020bottleneck} problem. This enables each graph node to be no more than one hop away from any other node. However, as a result, unnecessary information is shared between vectors that are not physically close but are part of the same polyline. Thus,~\cite{gao2020vectornet} only considers the nodes within a distance threshold to the predicted vehicle, in turn limiting the receptive field in aggregation. Furthermore, \cite{gao2020vectornet} includes ordering information into the node attributes via an integer index, which raises correctness questions since integers are combined with floating point features such as $xy$ positions.

Regarding the \textbf{map-representation} aspect of our work, instead of connecting vectors by their polyline membership, we ask the question, which parts of a polyline are the most relevant \textit{to an agent}? We task a \ac{GAT}~\cite{velivckovic2017graph} with the answer; the attention mechanism learns to determine the most relevant vectors without artificially limiting the receptive field. Furthermore, our map representation is simpler to pre-process than~\cite{liang2020learning}, \cite{zeng2021lanercnn} since we use standard graph convolutions, as well as~\cite{khandelwal2020if}, \cite{pan2020lane}, since we do not manually select the reference polyline and allow other map element types to be attended to as well. 

\subsection{Joint graph-based interaction modeling}
Social interaction in a driving scene can inherently be represented as a natural graph, where nodes are agents and edges model their (weighted) connections. Hence, virtually all recent state-of-the-art approaches use graph-based learned models such as \ac{GNN}s and~\ac{MHA}, which is related to the Transformer architecture\footnote{Incidentally, the basic Transformer layer is equivalent to the \ac{GNN} \ac{GAT} layer, in the case of multiple attention heads and a fully-connected underlying graph~\cite{hamilton2020graph}.}~\cite{vaswani2017attention}. For the sake of brevity, we limit our review to joint prediction works~\cite{tang2019multiple}, \cite{rhinehart2019precog}, \cite{mercat2020multi}, \cite{casas2020implicit}, \cite{cui2021lookout}, \cite{suo2021trafficsim}, \cite{li2021spatio}, \cite{mo2021heterogeneous}, \cite{ngiam2021scene}, \cite{scibior2021imagining}.

Among these approaches,~\cite{tang2019multiple}, \cite{rhinehart2019precog}, \cite{casas2020implicit}, \cite{cui2021lookout}, \cite{suo2021trafficsim}, \cite{li2021spatio} use deep generative models. Here, they capture the uncertainty within a scene into a set of latent variables and sample from the latent space. The major drawback of such approaches is the randomness of their outputs because they require sampling from a latent distribution during inference. Thus, likelihood estimation of the predicted trajectory distribution is difficult. This is exacerbated in the single-step prediction approaches~\cite{tang2019multiple}, \cite{rhinehart2019precog}, which construct a trajectory iteratively. Another drawback is limited map information; approaches either don't consider the map at all~\cite{mercat2020multi}, or use global representations centered around the \ac{AV}~\cite{casas2020implicit}, \cite{cui2021lookout}, \cite{suo2021trafficsim}. As mentioned in Sec.~\ref{sec:intro}, the \ac{AV}-centered global view requires extracting local patches around each agent of interest and relating information implicitly via \ac{CNN}s. As an alternative,~\cite{ngiam2021scene} and \cite{mo2021heterogeneous} perform joint prediction in the global reference frame without \ac{AV}-centering, but reduce the effects of arbitrary origin placement by injecting the global-frame map via cross-attention and \ac{LSTM}-like \cite{hochreiter1997long} implicit gating, respectively.

Regarding the \textbf{joint interaction modeling} aspect of our work, we start with deterministic one-shot prediction outputs given deterministic inputs. We model the map explicitly from local-frame graph representations with unlimited field of view. Our work is closest to the prediction model in~\cite{ngiam2021scene}, however, instead of using global reference frames, we ask the question, how is an agent jointly perceived by other agents? We arrive at a~\ac{MHA} model with masking, which combines multiple local representations to construct an implicit global frame. Furthermore, our model is smaller since it uses two \ac{GAT} and two~\ac{MHA} layers to model the map and social interaction, compared to 18~\ac{MHA} layers in~\cite{ngiam2021scene}. Finally, we frame the model in the action-space framework of our previous work~\cite{janjos2021action}, ensuring kinematically feasible predictions.


\section{METHOD}
\subsection{Background}

Consider the task of vehicle trajectory prediction in a driving situation with $N$ heterogeneous interacting agents. We define the single-agent prediction problem as predicting the distribution of future waypoints $\mathbf{\hat{Y}}_i$ of vehicle~$i$. It can be framed in an imitation learning setting, where a learned model parameterizes the distribution
\begin{equation}\label{eq:single_dist}
\hat{Y}_i \sim P(\mathbf{\hat{Y}}_i | \mathcal{D}^{i})\ ,
\end{equation}
conditioned on the local context $\mathcal{D}^i$ of vehicle $i$. Here, the superscript indicates that the values are represented in the local reference frame of vehicle $i$, subscript indicates prediction for agent $i$, while $\hat{\cdot}$ denotes predicted future values. We simplify the problem in \eqref{eq:single_dist} by predicting a sample $\hat{Y}_i$ of the distribution, e.g. a $2\times T$ matrix of $xy$ coordinates over $T$ future time steps.

The context $\mathcal{D}^i=\{\mathcal{M}^i, \mathcal{T}^i\}$ contains the map $\mathcal{M}^i$ and past position tracks $\mathcal{T}^i$ of vehicle $i$ and its neighboring $N-1$ agents, with $\mathcal{T}^i=\{X_j^i\}^N_{j=1}$. Each track $X_j^i$ is a $3\times T$ matrix of $xy$ coordinates of agent $j$ over $T$ past time steps (in the reference frame of vehicle $i$ at the current time step) and a padding row, since the agent might not be present in the scene for each time step. The padding row contains zeros for non-existent points (and zero positions) and ones otherwise.

In joint prediction, we consider the task of predicting $K$ ($K\leq N$) vehicles' trajectories simultaneously. Thus, the learned model parameterizes the distribution
\begin{equation}\label{eq:joint_dist}
\hat{Y} \sim P(\mathbf{\hat{Y}} | \mathcal{D})\ .
\end{equation}
Therefore, we predict a sample $\hat{Y}$ of $\mathbf{\hat{Y}}$ for future trajectories of all $K$ vehicles, given their contexts $\mathcal{D}$, where $\hat{Y}=\{\hat{Y}_k\}^K_{k=1}$ and $\mathcal{D}=\{\mathcal{D}^k\}^K_{k=1}$, respectively. Each $\mathcal{D}^k$ can be separated into its map and tracks components, $\mathcal{M}^k$ and $\mathcal{T}^k=\{X_j^k\}^N_{j=1}$. Note that tracks $\mathcal{T}^k$ contain trajectories for all $N$ agents, including those whose waypoints are not predicted such as pedestrians and bicycles.

In parameterizing the distributions in \eqref{eq:single_dist} and \eqref{eq:joint_dist}, we use a general encoder--decoder structure, with a context encoder and an output decoder. The encoder reasons about the context $\mathcal{D}$ while the decoder generates predicted trajectories $\hat{Y}$. Furthermore, we make use of track encoders, e.g. we encode each agent track $X_j^i$ ($i\in[1,...,K]$, $j\in[1,...,N]$) into a feature vector $z_j^i$ via a 1D \ac{CNN} network. 

\subsection{Action-space prediction}\label{subsec:asp}
We use positional representations in modeling the environment $\mathcal{D}$ within the context encoder. Both the polyline map $\mathcal{M}$ (as will be shown later), as well as the tracks $\mathcal{T}$, contain $xy$ coordinates describing polyline control points and past trajectories, respectively. Hence, in generating learned feature representations of the environment, map-dependent interactions are inferred from data in the Euclidean space.

However, when generating the predicted trajectories $\hat{Y}$ via the decoder, we shift the learning problem into the action-space of accelerations and steering angles. We provide past actions as action tracks $A_i^i$ ($i\in [1,...,K]$) and generate future actions with a~\ac{GRU} \cite{cho2014learning}. This is consistent with the action-space prediction framework~\cite{janjos2021action} and guarantees that a learned model does not have to capture motion models, as well as ensuring kinematic feasibility (with an output kinematic model). Similarly to position tracks, we encode past action tracks $A_i^i$ into feature vectors $w_i^i$ with the same network. 

\begin{figure}[]
	\hspace{-65pt}
	\scalebox{0.73}{\input{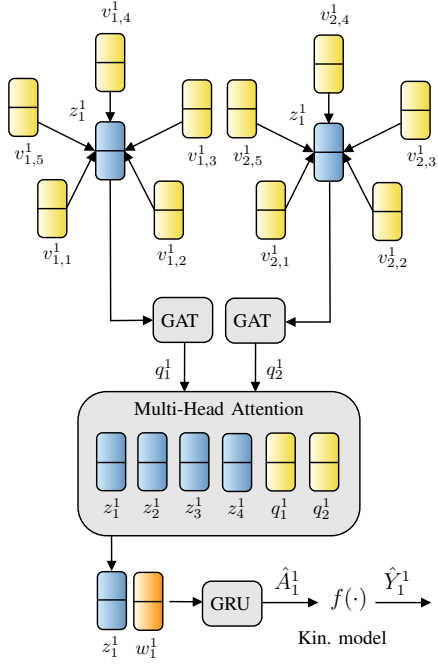}}
	\vspace{-3pt}
	\caption{\small StarNet single-agent architecture for the driving scene in Fig. \ref{fig:interaction_example}. Vehicle $1$ is the prediction-ego, and two polylines and three neighboring agents are in its local context. First, the two polylines determined by five vectors each are represented in a directed star graph with prediction-ego track embedding $z^1_1$ in the center and vector embeddings $v^1_{\{1,2\}, \{1,5\}}$ outward. Polyline-level embeddings are generated by a 2-layer \ac{GAT}, followed by a 1-layer Multi-Head Attention modeling map-dependent social interaction via aggregating all agents' and polylines' embeddings, $z^1_{\{1,4\}}$ and $q^1_{\{1,2\}}$ respectively. Then, prediction-ego embedding is selected, concatenated with its action track embedding $w^1_1$, and fed through an action-prediction \ac{GRU} to predict future actions $\hat{a}^1_1$. Finally, positions $\hat{x}^1_1$ are obtained via a kinematic model transformation.}
	\label{fig:starnet_block}
	\vspace{-15pt}
\end{figure}

\subsection{Single-agent prediction with StarNet}\label{subsec:starnet}
In this section, we present a single-agent prediction method for regressing the future trajectory of vehicle $i$ (prediction-ego). We approximate \eqref{eq:single_dist} via a deterministic encoder--decoder model. The encoder consists of a star graph map model and a map-dependent interaction model, while the decoder is a multi-modal action predictor, directly generating $m$ samples from the distribution \eqref{eq:single_dist}.

\subsubsection{Star graph map model}\label{subsec:stargraph}
Key component of single-agent StarNet is its representation of map elements via star graphs. First, each map element, such as a sidewalk, lane center-line, or a traffic island, is approximated by a polyline consisting of fixed-length vectors, similarly as in VectorNet~\cite{gao2020vectornet}. Thus, the representation $\mathcal{M}^i$ of the map in the agent $i$'s reference frame consists of $Q$ polylines
\vspace{-3pt}
\begin{equation}
\mathcal{M}^i = \{q^i_j\}^Q_{j=1}\ .
\vspace{-3pt}
\end{equation}In turn, each polyline consists of $L$ vectors comprising their start and end $xy$ coordinates and one-hot type encoding
\vspace{-3pt}
\begin{align}
q^i_j &= \{v^i_{jl}\}^L_{l=1}\ , \\
v^i_{jl} &= [v_{\text{start}}, v_{\text{end}}, v_{\text{type}}]^T\ . \label{eq:vector}
\end{align}

Given this polyline representation, we construct a directed star graph for each polyline $q^i_j$. In this structure, the past prediction-ego track is the central node with embedded track features $z^i_i$ and edges connecting each vector $v^i_{jl}$ to the central node. To ensure message passing compatibility, we embed the vector nodes \eqref{eq:vector} into the same dimensionality as the features $z^i_i$ using a linear layer. Finally, we feed this graph to a 2-layer \ac{GAT} and aggregate the nodes via max-pooling to obtain polyline-level embeddings $q^i_j$ of same dimensionality as the nodes. This structure is depicted in Fig. \ref{fig:starnet_block}.

The star graph and the accompanying \ac{GAT} model the relationship between the ego track and the map. Contrary to VectorNet's fully-connected graphs, we assume that there is more information contained in the vehicle's direct attention (represented by its past track embedding) to a specific vector within a polyline rather than between the polyline vectors themselves. This allows us to expand the receptive field, since the attention mechanism will learn to ignore distant vectors, and to consider them proportionally to their weights in aggregation. Furthermore, the structure removes the need to include artificial ordering into the vector \eqref{eq:vector}, as it is done in VectorNet in order to help convey the polyline geometry, which simplifies the learning problem.

\subsubsection{Map-dependent interaction model}\label{subsec:map_dependent_interaction}
In the single-agent StarNet, we model map-dependent social interaction with a \ac{MHA}~\cite{vaswani2017attention}, see Fig. \ref{fig:starnet_block}. We combine vehicle $i$'s past track embedding $z^i_i$ with polyline embeddings $q^i_{j}$ ($j\in[1,...,Q]$), as well as track embeddings $z^i_j$ ($j\in [1,...,N-1]$) of each agent sharing the scene with $i$. Then, we stack the embedding vectors into a matrix with $N+Q$ rows and feed it into a single \ac{MHA} layer. Here, linear projections of the input are generated in the form of query, key, and value matrices. Then, the self-attention operation~\cite{vaswani2017attention} is applied in order to infer the relationships between the embeddings. The output of the \ac{MHA} is of the same dimensionality as the stacked input matrix, and we select the row that corresponds to the vehicle $i$. Through the \ac{GAT} and \ac{MHA} layers, the obtained embedding is able to capture the map-dependent social interaction in the local context of the prediction-ego. The next step is feeding it into the action output decoder.

\begin{figure}[t!]
	\begin{subfigure}{0.48\textwidth}
		\hspace{-120pt}
		\scalebox{0.73}{\input{images/joint-starnet_block_diagram.tex}}
		\vspace{-3pt}
		\caption{\footnotesize Joint-StarNet with both map element star-graphs and action decoder blocks omitted: Single-agent Multi-Head Attention blocks model local map-dependent social interaction for the joint prediction candidate vehicles $1-3$. Features $z^1_{1-3}$, $z^2_{1-3}$, $z^3_{1-3}$ are selected and fed into the Joint Multi-Head Attention block, which aggregates local features to construct an implicit global frame. After this operation, features $z^i_i$ ($i\in[1,2,3]$) are selected and together with their action embeddings $w^i_i$ fed into the action decoder block (not shown).}
		\label{fig:joint-starnet_block}
	\end{subfigure}
	\begin{subfigure}{0.48\textwidth}
		\centering
		\scalebox{0.73}{\input{images/joint-starnet_mask.tex}}
		\caption{\footnotesize Depicted is the corresponding $9\times9$ attention mask matrix for the three vehicles from Fig \ref{fig:joint-starnet_block}. Here, each non-zero number denotes attention to a feature vector of a vehicle in a row, while zero denotes no attention. Each $3\times3$ matrix in a block-row can be obtained by left-shifting the left-neighbor $3\times3$ matrix by one.}
		\label{fig:joint-starnet_mask}
	\end{subfigure}
	\caption{\small Joint-StarNet architecture in Fig.~\ref{fig:joint-starnet_block} (for the scene in Fig.~\ref{fig:starnet_block}) and the accompanying attention mask for the Joint Multi-Head Attention block in Fig.~\ref{fig:joint-starnet_mask}.}
	\vspace{-19pt}
\end{figure}

\subsubsection{Multi-modal action decoder}
The action decoder combines the positional embedding of the prediction-ego, aggregated to consider the map-dependent social interaction, with its action embedding. We concatenate $z^i_i$ with $w^i_i$ and feed it into the action decoder, which is the same \ac{GRU} network as in~\cite{janjos2021action} (depicted in Fig. \ref{fig:starnet_block}). It generates steering angles and accelerations, directly predicting $m$ action modes (action trajectories and softmax scores, which can be interpreted as probabilities). The modes are converted to predicted vehicle positions via a bicycle kinematic model, fully capturing kinematic characteristics of motion.

We train the whole pipeline with the loss
\begin{equation}\label{eq:loss}
	\mathcal{L} = \mathcal{L}_{\text{reg}} + \beta\mathcal{L}_{\text{class}} \ ,
\end{equation}
where $\mathcal{L}_{\text{reg}}$ considers the mismatch to the ground-truth future trajectory and $\mathcal{L}_{\text{class}}$ considers the mode probability via cross-entropy, same as in~\cite{janjos2021action}, with $\beta$ set to $1$. 

\subsection{Joint prediction with joint-StarNet}\label{subsec:joint-starnet}
In this section, we present a joint prediction method for regressing the future trajectory of $k$ vehicles in a driving scene ($k\in[1,...,K], K\leq N$). We directly model the joint distribution \eqref{eq:joint_dist} without factorizing individual agents. The joint-StarNet architecture builds on StarNet from Sec.~\ref{subsec:starnet} and achieves joint prediction by aggregating local features into an implicit global frame via masked self-attention.

\subsubsection{Implicit global frame self-attention}
The joint-StarNet is an extension to single-agent StarNet. After determining the joint prediction candidates, we perform the first two steps of the single-agent StarNet pipeline separately for each vehicle. We construct local map element star graphs, aggregate them with \ac{GAT}s and combine them with locally embedded tracks in the Single-Agent \ac{MHA} blocks. Then, we select the positional embeddings corresponding to each of the joint agents, in each joint agent's local context, obtaining $K^2$ feature vectors~$\{\{z^k_j\}^K_{j=1}\}^K_{k=1}$. An example is provided in Fig. \ref{fig:joint-starnet_block}. This combination of features contains mutual local information about each joint prediction candidate, at the cost of quadratically increasing number of features. Nevertheless, we do not observe a computational bottleneck due to efficient batching in training, described in Sec.~\ref{sec:implementation}.

Given the individual local features, we now construct an implicit global frame by combining features from each local frame. We achieve this with another \ac{MHA} block (denoted as Joint Multi-Head Attention in Fig. \ref{fig:joint-starnet_block}), taking in features~$\{\{z^k_j\}^K_{j=1}\}^K_{k=1}$ stacked into a matrix. In the output, we select the rows corresponding to features $\{z^k_k\}^K_{k=1}$ (in their respective reference frames) and feed them in a batched manner into an action decoder block. We can train with the same loss \eqref{eq:loss} as in single-agent StarNet training.

\subsubsection{Attention-mask}\label{subsubsec:joint_mask}
In combining multiple local contexts into an implicit global context, the embeddings corresponding to a single vehicle in different local frames should attend only to themselves. We achieve this by limiting the self-attention with a $K^2\times K^2$ attention mask matrix. It ensures that only the features~$\{z^k_j\}^K_{k=1}$ for the agent $j$ in different frames are considered, in each row of the stacked input matrix. This is exemplified in Fig.~\ref{fig:joint-starnet_mask}. 

The joint-StarNet architecture with masking allows to explicitly combine multiple local interaction models and integrate them into an implicit global interaction model while accounting for non-symmetric attention. Each local, single-agent model uses direct map representations that condition the local social interaction. 

\begin{table}[]
	\centering
	\smallskip
	\scalebox{0.88}{\begin{tabularx}{0.5\textwidth}{c *{4}{Y}}
	\toprule
	& \multicolumn{2}{c}{inD \cite{bock2019ind}}  
	& \multicolumn{2}{c}{INTERACTION \cite{zhan2019interaction}}\\
	\cmidrule(lr){2-3} \cmidrule(l){4-5}
	& ADE & FDE & ADE & FDE\\
	\midrule
	FFW-ASP \cite{janjos2021action}  & 0.37 &  1.02 & 0.24 & 0.63 \\
	VectorNet \cite{gao2020vectornet}  &  0.37 & 1.03 & 0.22 & 0.63 \\
	\cmidrule(lr){2-3} \cmidrule(l){4-5}
	StarNet  &  0.35 & 0.97 & 0.16 &  0.49 \\
	joint-StarNet\textit{-no-mask}  &  0.36 & 1.02 & 0.15 & 0.43 \\
	joint-StarNet  &  \textbf{0.32} & \textbf{0.89} & \textbf{0.13} & \textbf{0.38} \\
	\bottomrule
\end{tabularx}}
	\caption{\small Comparison of StarNet and joint-StarNet (also without masking from Sec.~\ref{subsubsec:joint_mask}) with FFW-ASP~\cite{janjos2021action} and VectorNet~\cite{gao2020vectornet} (own implementation), taking the best out of $m=3$ modes.}
	\label{tab:results_reproduced}
	\vspace{-11pt}
\end{table}

\begin{table}[]
	\centering
	\smallskip
	\scalebox{0.88}{\begin{tabularx}{0.5\textwidth}{c *{2}{Y}}
	\toprule
	& \multicolumn{2}{c}{INTERACTION \cite{zhan2019interaction}}  \\
	\cmidrule(l){2-3}
	& ADE & FDE\\
	\midrule
	DESIRE \cite{lee2017desire}  &  0.32 & 0.88 \\
	MultiPath \cite{chai2019multipath}  &  0.30 & 0.99 \\
	STG-DAT \cite{li2021spatio} & 0.29 & 0.54 \\
	TNT \cite{zhao2020tnt}  &  0.21 & 0.67 \\
	ReCoG \cite{mo2020recog}  &  0.19 & 0.66 \\
	HEAT-I-R \cite{mo2021heterogeneous}  &  0.19 & 0.65 \\
	ITRA \cite{scibior2021imagining}  &  0.17 & 0.49 \\
	\cmidrule(l){2-3}
	StarNet  &  0.16 & 0.49 \\
	joint-StarNet  & \textbf{0.13} & \textbf{0.38} \\
	\bottomrule
\end{tabularx}}
	\caption{\small Comparison of StarNet and joint-StarNet with approaches in literature (reported results). The values for~\cite{lee2017desire} and~\cite{chai2019multipath} are given in~\cite{zhao2020tnt}. Since~\cite{li2021spatio} reports results for different map types separately, we computed the aggregate value by combining the ratios of specific map types in the validation dataset.}
	\label{tab:results_literature}
	\vspace{-15pt}
\end{table}
\vspace{-2pt}
\section{RESULTS}
\vspace{-2pt}
\subsection{Implementation}\label{sec:implementation}
For implementing the StarNet (Sec.~\ref{subsec:starnet}) and joint-StarNet (Sec.~\ref{subsec:joint-starnet}) architectures we used several network types: 1D \ac{CNN}s, \ac{GAT}s, \ac{MHA}, and \ac{GRU}s. The track encoder 1D \ac{CNN}s are adapted from the ActorNet model in~\cite{liang2020learning} and embed position and action tracks to 128- and 64-dimensional vectors, respectively. The position embeddings are then used as nodes in the \ac{GAT} and \ac{MHA} networks, which are both realized with 8 attention heads. In the action decoder block, the concatenated position and action track embeddings are first transformed by two linear layers of sizes \{512, 256\} (with batch normalization and $\tanh$ activation) before being fed into the \ac{GRU} network. The \ac{GRU} iterates these transformed features three times through two layers of 512 hidden units, directly predicting $m$ future action modes. 

The joint-StarNet has several important practical considerations. Within the model, each joint candidate is first fed through the first two steps of the single-agent StarNet (Sec. \ref{subsec:stargraph} and Sec. \ref{subsec:map_dependent_interaction}), and then the features from multiple local contexts are aggregated in the Joint \ac{MHA}, as exemplified in Fig.~\ref{fig:joint-starnet_block}. In a single batch element, this induces linearly growing complexity in the GAT and Single-agent \ac{MHA} and quadratically growing complexity in the Joint \ac{MHA}, with the number of joint candidates. However, the computational load does not grow equally due to efficient batching of different scenes. In the GAT case, we aggregate different star graphs into a single graph with a block-diagonal adjacency matrix. Similarly, in both \ac{MHA} blocks we feed inputs from different scenes together in a batch, but use additional batch-wise attention mask. As a result, this brings a higher utilization of GPU memory. However, reasonably-sized batches are made possible by the compact input representations.

\begin{figure*}[h!]
	\centering
	\begin{subfigure}[b]{0.4\columnwidth}
		\frame{\includegraphics[width=\linewidth]{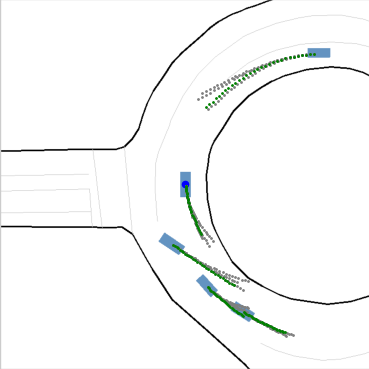}}
	\end{subfigure}
	\begin{subfigure}[b]{0.4\columnwidth}
		\frame{\includegraphics[width=\linewidth]{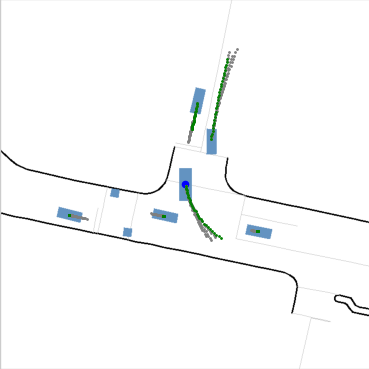}}
	\end{subfigure}
	\begin{subfigure}[b]{0.4\columnwidth}
		\frame{\includegraphics[trim=0 0 0 30, clip,width=\linewidth]{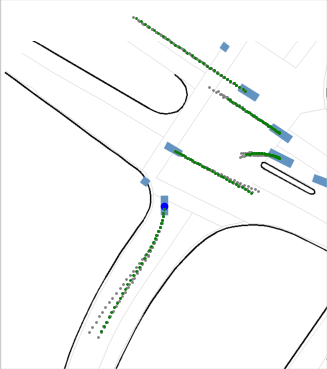}}
	\end{subfigure}
	\begin{subfigure}[b]{0.4\columnwidth}
		\frame{\includegraphics[trim=0 0 0 16, clip,width=\linewidth]{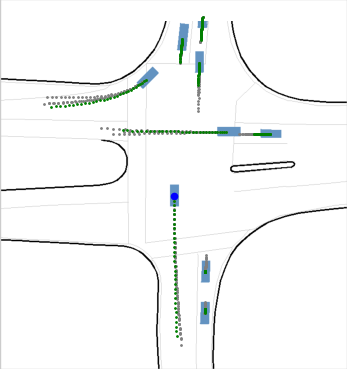}}
	\end{subfigure}
	\begin{subfigure}[b]{0.4\columnwidth}
		\frame{\includegraphics[width=\linewidth]{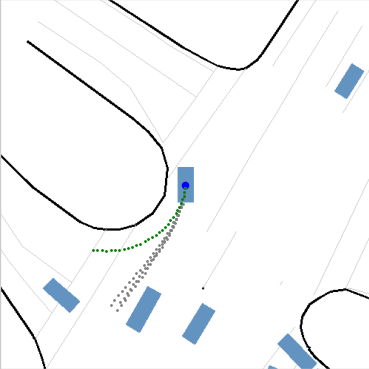}}
	\end{subfigure}
	\caption{Qualitative results of joint-StarNet on INTERACTION~\cite{zhan2019interaction} validation subset. Rectangles indicate vehicles, while squares are pedestrians/bicycles. Gray trajectories are predictions ($m=3$ modes), with the ground truth overlaid in green. The right-most prediction is among the 50 worst individual predictions according to the FDE metric.}
	\label{fig:example_predictions}
	\vspace{-20pt}
\end{figure*}

\subsection{Datasets}
Datasets must allow flexibility in choosing single or multiple prediction targets, in order to facilitate both single-agent and joint trajectory prediction. Therefore, we selected the inD~\cite{bock2019ind} and the INTERACTION datasets~\cite{zhan2019interaction} that provide joint tracked data over a whole recording. In both datasets, we generated individual samples for all agents except pedestrians and bicycles by extracting all 2.5+3\si{s} segments (2.5\si{s} past and 3\si{s} prediction recorded with 10\si{Hz}) with a 1.5\si{s} spacing between the samples. In generating map data, we used the provided lanelets~\cite{poggenhans2018lanelet2} to extract two polyline types: road boundaries (e.g. curbstones) and road properties (e.g. lane center-lines). For inD, we used the same training/validation/testing split as in~\cite{janjos2021action}, and for INTERACTION we used the provided validation dataset.

In the joint-StarNet setting, it is non-trivial to determine the joint prediction candidates. We perform this by first selecting one vehicle in the scene that is present over an entire prediction time interval as the virtual-ego. Then, we determine all other vehicles that exist throughout the time interval and that at any time during the interval come close to the virtual-ego within a certain threshold; better approaches to filter relevant agents are left for future work. In this sense, the virtual-ego mimics an \ac{AV} that predicts trajectories of its nearby vehicles (and its own trajectory). Thus, when another vehicle is chosen to be the virtual-ego, a different set of joint candidates could emerge. This allows us to effectively upsample highly interactive training data.

\subsection{Training setup}
We implemented our models in PyTorch \cite{paszke2019pytorch} and trained with the Adam optimizer \cite{kingma2014adam} over~10 epochs with batch size~8. The learning rate was set to~$10^{-4}$ and multiplied by~$0.2$ for every two consecutive epochs with no improvement outside an $\epsilon$ region. For joint-StarNet, the training took around five days to complete on a single RTX 3090 GPU.

\subsection{Performance}\label{subsec:performance}
We compared our approaches against the raster-based \ac{FFW-ASP} architecture from our previous work~\cite{janjos2021action} and our own implementation of VectorNet~\cite{gao2020vectornet}, with the results shown in Tab.~\ref{tab:results_reproduced}. To ensure a fair comparison, we used our multi-modal action output decoder in VectorNet (originally uni-modal) and did not incorporate the self-supervised map completion task from~\cite{gao2020vectornet} within the \ac{MHA} blocks of any architecture. Similarly, we adapted the \ac{FFW-ASP} to use action track encoders mentioned in Sec. \ref{subsec:asp}. As metrics, we computed the~\ac{ADE} and~\ac{FDE}, defined as in~\cite{scibior2021imagining}. 

As seen in Tab. \ref{tab:results_reproduced}, StarNet's graph-based map modeling and map-conditioned interaction modeling already brings improvements over the baseline methods. Similarly, joint-StarNet improves on StarNet's performance, as expected. Tab. \ref{tab:results_reproduced} can also be interpreted as an ablation study, where the effects of removing the graph-based map modeling (by \ac{FFW-ASP}/StarNet results) and masked joint prediction (by StarNet/joint-StarNet(\textit{-no-mask}) results). We also compared against reported results in literature on the INTERACTION validation dataset, see Tab.~\ref{tab:results_literature}. To the best of our knowledge, joint-StarNet achieves state-of-the-art performance. 

Examples of predicted trajectories with joint-StarNet are shown in Fig.~\ref{fig:example_predictions}. It can be seen that overall the predictions accurately model the interaction in a scene. Regarding multi-modality, we observe that in most cases the modes are distributed realistically, i.e. in straight driving they are spread longitudinally and slight turns result in relatively narrow dispersion. However, among the shown worst prediction example, turns with a larger radius result in occasional missed modes, indicating room for improvement.

\vspace{-1pt}
\section{CONCLUSION}
\vspace{-1pt}
In this work, we presented an attention-based approach to directly represent map elements and explicitly model mutual social interaction. We offered two novel architectures, the single-agent StarNet that models map elements as star graphs, and a joint prediction extension via an additional \ac{MHA} layer. The joint-StarNet can handle a variable number of agents and integrate their local awarenesses into an implicit global model. In this sense, it contributes an important step towards joint scene understanding. 

In future work, we will focus on the multi-modality aspect of joint prediction and address the shortcomings mentioned in Sec.~\ref{subsec:performance}. We plan to improve on the implicit modeling of multi-modality within the action output decoder, which does not condition predicted modes on other vehicles' predicted modes. Furthermore, we plan to integrate the presented architectures with the self-supervised long-term prediction framework of~\cite{janjos2021action}, which predicts future context representations prior to trajectories. We expect that the denser map and joint interaction modeling will lead to improved context prediction, bringing further performance improvements.

\newpage
\bibliographystyle{IEEEtran}
\bibliography{references/bibliography}

\end{document}